%% file: main.tex
\documentclass{l4dc2024}

\usepackage{booktabs}
\usepackage{multirow}
\usepackage{siunitx}

\title[]{Risk-Sensitive Stochastic Optimal Control as \\ Rao-Blackwellized Markovian Score Climbing}

\usepackage{mathtools}
\usepackage{times}
\usepackage{tikz}
\usetikzlibrary{plotmarks}

\usepackage{wrapfig}
\usepackage[
    labelfont=bf,
    justification=justified,
    format=plain
]{caption}

\usepackage{pgfplots}
\pgfplotsset{compat=1.11}

\newcommand{\given}{\,|\,}

\newcommand{\E}{\mathbb{E}}
\newcommand{\dd}[0]{\mathrm{d}}
\DeclareMathOperator*{\argmin}{arg\,min}
\DeclareMathOperator*{\argmax}{arg\,max}

\coltauthor{%
 \Name{Hany Abdulsamad} \Email{hany.abdulsamad@aalto.fi} \\
 \Name{Sahel Iqbal} \Email{sahel.iqbal@aalto.fi} \\
 \Name{Adrien Corenflos} \Email{adrien.corenflos@aalto.fi} \\
 \Name{Simo S\"arkk\"a} \Email{simo.sarkka@aalto.fi} \\
 \addr Aalto University, Espoo, Finland
}

\begin{document}

\maketitle

\begin{abstract}
    Stochastic optimal control of dynamical systems is a crucial challenge in sequential decision-making. Recently, control-as-inference approaches have had considerable success, providing a viable risk-sensitive framework to address the exploration-exploitation dilemma. Nonetheless, a majority of these techniques only invoke the inference-control duality to derive a modified risk objective that is then addressed within a reinforcement learning framework. 
    This paper introduces a novel perspective by framing risk-sensitive stochastic control as Markovian score climbing under samples drawn from a conditional particle filter.
    Our approach, while purely inference-centric, provides asymptotically unbiased estimates for gradient-based policy optimization with optimal importance weighting and no explicit value function learning. To validate our methodology, we apply it to the task of learning neural non-Gaussian feedback policies, showcasing its efficacy on numerical benchmarks of stochastic dynamical systems.
\end{abstract}


\section{Introduction}\label{sec:intro}
Stochastic optimal control serves as a fundamental paradigm of low-level decision-making under uncertainty, with applications ranging from chemical plant control to autonomous driving. Current methods in this domain can be broadly categorized into those that rely on analytical modeling and local approximations~\citep{rakovic2018handbook} and others that fall under a data-driven stochastic optimization class~\citep{deisenroth2011pilco, levine2013guided}. 

Recently, the success of data-driven methods in tackling complex problems has led to their increasing prominence. Many of these approaches can be interpreted through the lens of risk-sensitive optimization~\citep{jacobson1973optimal, whittle1990risk}, effectively balancing exploration-exploitation dynamics. Conveniently, the paradigm of risk-sensitive decision-making elegantly showcases the well-known duality between control and inference~\citep{kalman1960new, stengel1986stochastic, attias2003planning, toussaint2006probabilistic, todorov2008general, kappen2012optimal}. However, despite acknowledging this connection in formulating a risk-sensitive objective, a substantial portion of existing research tends to revert to conventional paradigms in control and reinforcement learning~\citep{levine2018reinforcement}, neglecting the vast toolbox offered by modern Bayesian inference techniques~\citep{martin2022computing}. 

This paper introduces a novel approach to stochastic nonlinear control embedded within an inference framework, capitalizing on state-of-the-art Markov chain Monte Carlo~\citep{brooks2011handbook} and sequential Monte Carlo~\citep{chopin2020intro} techniques. Early control-as-inference literature primarily focused on tractable settings, either in discrete or linear-Gaussian environments~\citep{attias2003planning, todorov2008general, toussaint2006probabilistic}. While these methods were later adapted for nonlinear settings via belief propagation~\citep{toussaint2009robot, rawlik2013probabilistic} and iterated smoothing~\citep{watson2020stochastic}, they struggle especially in highly nonlinear, non-Gaussian, and bounded domains.

More recent methods avoid these pitfalls by employing particle-based techniques~\citep{williams2017model, pinneri2021sample}. However, these methods often only involve pure forward propagation, intermediate Gaussian approximations, heuristic particle clipping and heuristic particle rejuvenation techniques. Even certain variants that utilize particle filters~\citep{piche2018probabilistic} do so in combination with state-value function approximations, thus getting entangled in the reinforcement learning paradigm and neglecting the rich literature on particle smoothing techniques. %

Our approach distinguishes itself by steering clear of Gaussian approximations and heuristic methods, aiming for a more direct and rigorous treatment of the problem via Monte Carlo simulation that does not incur bias. By invoking the duality between control and inference, we formulate the problem of optimal decision-making as policy learning in an equivalent nonlinear and non-Gaussian state-space model. We learn these policies via a Markovian score climbing technique~\citep{gu1998stochastic, naesseth2020markovian} under Monte Carlo samples. These are simulated by a Rao-Blackwellized Markov chain whose target density is the optimal state-action trajectory distribution of the corresponding risk-sensitive stochastic control problem.

The paper is structured as follows. In \sectionref{sec:background}, we briefly introduce the problem of stochastic optimal control and revisit the duality between control and inference by framing the risk-sensitive control objective as the marginal likelihood of an equivalent state-space model. In \sectionref{sec:csmc_soc}, we propose a novel policy learning technique based on optimizing the log marginal likelihood objective. More concretely, we leverage the state-space interpretation of risk-sensitive stochastic control to formulate a Rao-Blackwellized conditional sequential Monte Carlo (RB-CSMC) kernel which provides low variance, unbiased estimates of the derivative of the log marginal likelihood, and by proxy the control objective. Finally, in \sectionref{sec:experiments}, we apply our method on numerical examples to learn neural feedback policies in control scenarios of stochastic dynamical systems.

\section{Background}\label{sec:background}
This section is organized as follows: in Section~\ref{sec:soc}, we first introduce the problem statement of stochastic control; and in Section~\ref{sec:risk_soc}, we provide a brief presentation of the control-inference duality this paper relies on.

\subsection{Stochastic Optimal Control}\label{sec:soc}
Consider a discrete-time finite-horizon decision process characterized by a state space $\mathcal{X} \subseteq \mathbb{R}^{d}$ and an action space $\mathcal{U} \subseteq \mathbb{R}^{m}$. Given a non-negative cost function $c(x_t, u_t)$, we formulate the expected total cost criterion~\citep{puterman2014markov} as
\begin{equation}\label{eq:soc_objective}
  \mathcal{J}(\pi) = \E \left[\sum_{t=0}^{T} c(x_t, u_t)\right],
\end{equation}
where the expectation is computed under the state-action trajectory distribution
\begin{equation}
    p(x_{0:T}, u_{0:T}) = \mu(x_{0}) \prod_{t=0}^{T-1} f(x_{t+1} \given x_t, u_t) \, \prod_{t=0}^T \pi(u_t \given x_t),
\end{equation}
with an initial state distribution $\mu(x_{0})$, stochastic dynamics $f(x_{t+1} \given x_t, u_t)$, and a stochastic policy $\pi(u_{t} \given x_{t})$. The initial state $x_0 \in \mathcal{X}$ is often known, in which case $\mu(x_{0}) = \delta(x_{0})$, where $\delta$ is the Dirac delta distribution. For a policy space $\Pi$, the stochastic optimal control problem solves
\begin{equation}
  \pi^{*} = \argmin_{\pi \in \Pi} \medspace \mathcal{J}(\pi).
\end{equation}

In practice, we choose a parametric form for the stochastic policy, $\pi(\cdot \given x_t) \equiv \pi_{\theta}(\cdot \given x_t)$, with a parameter vector $\theta \in \Theta \subseteq \mathbb{R}^{l}$. Consequently, the expected cost objective becomes a function of $\theta$, and the goal is to identify a vector $\theta^*$ that minimizes $\mathcal{J}(\theta)$. Note that, in a finite-horizon total-cost scenario, the optimal policy is in fact non-stationary. However, as the time horizon $T$ becomes sufficiently large, the problem can be regarded as an average-cost setting for which the optimal policy is stationary \citep{puterman2014markov}. Similar assumptions were made in the seminal work of \citet{deisenroth2011pilco} and \citet{levine2013guided}.

\subsection{Risk-Sensitive Stochastic Control as Inference}\label{sec:risk_soc}
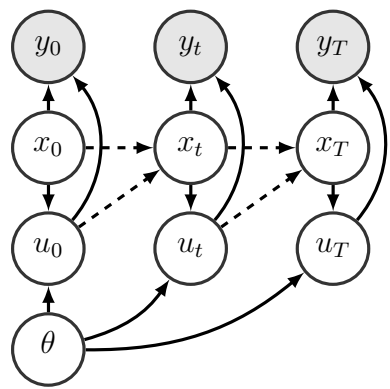
\begin{wrapfigure}{r}{0.4\textwidth}
    \centering
    \vspace{-1.2cm}
    \input{figures/dgm.tex}
    \vspace{-0.75cm}
    \caption{The finite-horizon graphical model underlying the control problem.}
    \label{fig:dgm}
\end{wrapfigure}

To transform the optimization objective from \sectionref{sec:soc} into an inference problem, we adopt the approach of \citet{toussaint2009robot} and \citet{rawlik2013probabilistic}. It consists in introducing a binary pseudo-measurement variable $y_t$ with a likelihood function defined as $$g(y_t = 1 \given x_t, u_t) = \exp \big\{-\eta \, c(x_t, u_t)\big\}, \quad \eta \in \mathbb{R}^+,$$ resembling a Boltzmann distribution. Here, $c(x_{t}, u_{t})$ serves as an energy function, and $\eta$ acts as the inverse temperature. This likelihood represents the probability that a given state-action pair $(x_t, u_t)$ is optimal \citep{dayan1997using, toussaint2006probabilistic}. For an overview on the role of $\eta$ in various control-as-inference algorithms, see \citet{watson2023inferring}.

This reinterpretation enables the formulation of a state-space model (SSM) aligned with the decision-making process as depicted in \figureref{fig:dgm}. The joint density over all random variables in this directed graph is expressed as
\begin{equation}
    \label{eq:ssm_elaborate}
    p(x_{0:T}, u_{0:T}, y_{0:T} \given \theta) = \delta(x_0) \prod_{t=0}^{T-1} f(x_{t+1} \given x_t, u_t) \, \prod_{t=0}^T \pi_{\theta}(u_t \given x_t) \, \prod_{t=0}^T g(y_t \given x_t, u_t).
\end{equation}
As demonstrated by \citet{toussaint2009robot}, the negative log marginal likelihood of this state-space model serves as a lower bound on the expected cost objective outlined in \equationref{eq:soc_objective}. Consequently, a direct equivalence between maximizing the marginal likelihood and minimizing the expected cost does not hold. Instead, the maximization of the log marginal likelihood amounts to the minimization of a risk-sensitive variant of the true objective:
\begin{equation}\label{eq:max_likelihood}
    \argmax_{\theta \in \Theta} \medspace \log p(y_{0:T} = 1 \given \theta) = \argmin_{\theta \in \Theta} \medspace -\frac{1}{\eta}\log \E_{p_{\theta}} \left[ \exp \big\{-\eta \textstyle \sum_{t=0}^{T} c(x_{t}, u_{t}) \big\} \right],
\end{equation}
where $p_{\theta} \coloneqq p(x_{0:T}, u_{0:T} \given \theta)$. Note that $\eta$ modulates the risk trade-off of the objective and that the risk-neutral objective is retrieved in the limit $\eta \rightarrow 0$~\citep{whittle1990risk}. To simplify notation, let us denote by $z_t$ the state-action pair $(x_t, u_t)$. Additionally, define $\mu_{\theta}(z_0) \coloneqq \delta(x_0) \, \pi_{\theta}(u_0 \given x_0)$ and $f_{\theta}(z_{t+1} \given z_t) \coloneqq f(x_{t+1} \given x_t, u_t) \, \pi_{\theta}(u_{t+1} \given x_{t+1})$. Then \equationref{eq:ssm_elaborate} can be written as
\begin{equation}
    \label{eq:ssm}
    p_{\theta}(z_{0:T}, y_{0:T}) = \mu_{\theta}(z_0) \prod_{t=0}^{T-1} f_{\theta}(z_{t+1} \given z_t) \prod_{t=0}^T g(y_t \given z_t).
\end{equation}
The joint density in \equationref{eq:ssm} defines an equivalent state-space model whose states are the state-action pairs $z_t$ and measurements are the optimality indicators $y_t$. Thus we see that the problem of (risk-sensitive) stochastic optimal control has now been reduced to the problem of maximum likelihood estimation in an SSM with fixed observations.

\section{Risk-Sensitive Stochastic Control as Markovian Score Climbing}\label{sec:csmc_soc}
We have reviewed the connection of risk-sensitive stochastic control to maximum likelihood optimization in \sectionref{sec:risk_soc}. While this perspective has been previously leveraged, related  approaches have primarily focused either on simple tractable settings \citep{toussaint2006probabilistic, hoffman2009expectation}, or relied on biased approximate methods \citep{toussaint2009robot, rawlik2012stochastic, watson2020stochastic}. In the following, we present our method that builds on state-of-the-art Markov chain Monte Carlo (MCMC) and sequential Monte Carlo (SMC) techniques to retrieve unbiased optima of the marginal likelihood and the risk-sensitive objective. The advantage of this approach lies in its ability to deal with the challenges of the stochastic control problem in nonlinear, non-Gaussian, and bounded domains, without resorting to the approximations and heuristics previously mentioned.

\subsection{Policy Optimization via Score Climbing}
Our policy learning procedure aims to maximize the log marginal likelihood $\ell(\theta) \coloneqq \log p(y_{0:T} = 1 \given \theta)$ in \equationref{eq:max_likelihood} using gradient ascent. Such an approach requires being able to compute the gradient of the log-likelihood, $\nabla_\theta \ell(\theta)$, also known as the \emph{score function}. When we have access to \emph{unbiased} estimators of the score function, we can use a stochastic gradient ascent algorithm, wherein given an initial guess $\theta_0 \in \Theta$ and a step size sequence $\{\gamma_k \given \gamma_k \in \mathbb{R}^+, k=1, \dots\}$, at iteration $k$, we update the parameters according to $\theta_k = \theta_{k-1} + \gamma_k \, \widehat{\nabla_\theta \ell}(\theta_{k-1})$. Here $\widehat{\nabla_\theta \ell}(\theta_{k-1})$ is a stochastic estimate of the score function evaluated at $\theta = \theta_{k-1}$. If the step sizes satisfy the conditions $\sum_{k=1}^\infty \gamma_k = \infty$ and $\sum_{k=1}^\infty \gamma_k^2 < \infty$, and if $\mathbb{E}\left[\widehat{\nabla_\theta \ell}(\theta_{k-1})\right] = \nabla_\theta \ell(\theta_{k-1})$, then this ascent converges to a local optimum~\citep{robbins1951stochastic}.

Unbiased estimates of the marginal likelihood can be retrieved with a particle filter. However, computing their gradients is challenging due to the non-differentiability of resampling schemes~\citep[see, e.g.,][]{maddison2017filtering}. While differentiable resampling schemes~\citep{corenflos2021differentiable} have been recently developed, they come at a substantial computational complexity and are therefore more often used to learn proposal distributions of particle filters, rather than parameters of the underlying model. We shall circumvent this difficulty altogether by making use of Fisher's identity~\citep{cappe2005inference},
\begin{equation}
    \label{eq:fishers_identity}
    \nabla_\theta \, \ell(\theta) = \int \nabla_\theta \log p_{\theta}(z_{0:T}, y_{0:T}) \, p_\theta(z_{0:T} \given y_{0:T}) \, \dd z_{0:T}.
\end{equation}
This identity yields a score estimate given a differentiable complete data likelihood $p_{\theta}(z_{0:T}, y_{0:T})$ and Monte Carlo samples from the smoothing distribution $p_\theta(z_{0:T} \given y_{0:T})$. In the context of SSMs, this is typically achieved using particle smoothing~\citep{kitagawa1996monte,godsill2004monte,poyiadjis2011particle}, and the resulting algorithm can be understood as an approximate expectation-maximisation routine~\citep{kantas2015particle}. %

\subsection{Markovian Score Climbing for Maximum Likelihood Estimates}
To compute the score in~\equationref{eq:fishers_identity}, one could use particle smoothers such as the forward filtering backward sampling algorithm~\citep[FFBS,][]{godsill2004monte} to generate samples from the smoothing posterior. However, particle smoothers provide biased estimates of expectations under the smoothing distribution for a finite number of samples $N$~\citep[Chapter 12]{kantas2015particle,chopin2020intro}. It is worth noting that while approximation errors are unavoidable due to numerical integration, this incurred bias in the score estimate adds to the approximation of the gradient descent in a way that is, in essence, uncontrollable since the \emph{true} score is unknown. Instead, we wish to use a method that is consistent, in the single limit of an infinite number of gradient steps, no matter the choice of $N$.

The stochastic gradient ascent using Markov chain Monte Carlo algorithm from \citet[reproduced in \algorithmref{alg:msc}]{gu1998stochastic} provides such a guarantee. This algorithm, later adapted by \citet{naesseth2020markovian} for variational inference under the name Markovian score climbing (MSC), relies on having access to a $p_\theta(z_{0:T} \given y_{0:T})$-ergodic Markov chain Monte Carlo (MCMC) kernel $\mathcal{K}(\cdot \given z_{0:T}, \theta)$. Formally, this means that applying $Z^n \sim \mathcal{K}(\cdot \given \mathcal{K}(\cdot \given \ldots, \theta), \theta) \eqqcolon \mathcal{K}^n(\cdot \given z_{0:T}, \theta)$ repeatedly on an arbitrary initial trajectory asymptotically produces samples from $p_\theta(z_{0:T} \given y_{0:T})$: the law of $Z^n$ converges to $p_\theta(z_{0:T} \given y_{0:T})$ and we have a law of large numbers
\begin{equation}\label{eq:lln}
    \frac{1}{N} \sum_{n=1}^N h(Z^n) \underset{N\to \infty}{\to} \int p_{\theta}(z_{0:T} \given y_{0:T}) \, h(z_{0:T}) \, \dd z_{0:T}.
\end{equation}
We refer to~\citet[Chapters 2 and 5]{douc2018markov} for more details on MCMC, the law of large numbers for Markov chains, and the conditions under which they define ergodic chains.

\begin{algorithm2e}[t]
    \caption{Markovian score climbing for maximum likelihood estimation}
    \label{alg:msc}
    \SetAlgoLined
    \LinesNumbered
    \DontPrintSemicolon
    \KwIn{Initial trajectory $z^{0}_{0:T}$, initial parameters $\theta_0$, number of iterations $M$, step sizes $\gamma_{1:M}$, Markov kernel $\mathcal{K}$.}
    \KwOut{$\theta^* \approx \theta_{\text{MLE}}$}
    \For{$k \gets 1,\dots, M$}{
        Sample $z^{k}_{0:T} \sim \mathcal{K}(\cdot \given z^{k-1}_{0:T}, \theta_{k-1})$ \;
        Compute $\widehat{\nabla_\theta \ell}(\theta_{k-1}) \gets \nabla_\theta \log p_\theta(z^{k}_{0:T}, y_{0:T}) \vert_{\theta=\theta_{k-1}}$ \;
        Update $\theta_k \gets \theta_{k-1} + \gamma_k \widehat{\nabla_\theta \ell}(\theta_{k-1})$\;
    }
    \Return{$\theta_M$}
\end{algorithm2e}

We defer the question of how to construct such a kernel for our SSM in~\equationref{eq:ssm} to \sectionref{subsec:csmc} and for now let us assume we have defined one. The MSC algorithm proceeds by sampling a trajectory from the Markov chain~(line 2 in \algorithmref{alg:msc}), computing the Monte Carlo estimate to the score under this one sample, and updating the policy parameters. It can be shown~\citep[Theorem 1]{gu1998stochastic} that under some technical regularity conditions on the target distribution and the associated Markov chain transition $\mathcal{K}(\cdot \mid z_{0:T}, \theta)$, including ergodicity, Algorithm~\ref{alg:msc} converges to the true maximum likelihood estimate. \citet{gu1998stochastic} used multiple samples from the Markov chain to compute the score at each iteration and incorporated second-order gradient information. However, as proved by \citet{naesseth2020markovian}, neither is necessary to ensure convergence.

\subsection{The Conditional Sequential Monte Carlo Kernel}\label{subsec:csmc}
Conditional sequential Monte Carlo~\citep[CSMC,][]{andrieu2010particle} is an MCMC kernel that targets the distribution $p_{\theta}(z_{0:T} \given y_{0:T})$ as requested. While its analysis is complicated, it can be implemented as a simple modification of a standard particle smoothing algorithm. Given a \textit{reference trajectory} $z_{0:T}$ from the smoothing (or \textit{target}) distribution, at each time step in the forward pass, $N-1$ particles are sampled conditionally on the reference particle surviving the resampling step~(see lines 6-10 in \algorithmref{alg:csmc}). After the filtering pass is complete, backward sampling (lines 12-15) is used to return a new reference trajectory $z^{*}_{0:T}$. This algorithm can be shown to be well-behaved (i.e., the chain it defines converges geometrically fast to the target) for as little as $N=2$ particles~\citep{chopin2015particle, andrieu2015uniform, lee2020coupled}. In~\algorithmref{alg:csmc}, we explicitly describe the CSMC kernel with backward sampling~\citep{andrieu2010particle, whiteley2010discussion} and multinomial resampling.
\begin{algorithm2e}[t]
    \caption{(Bootstrap) Conditional SMC with backward sampling}
    \label{alg:csmc}
    \SetAlgoLined
    \LinesNumbered
    \DontPrintSemicolon
    \SetKwInput{Notation}{Notation}
    \KwIn{Trajectory $z_{0:T}$}
    \KwOut{New reference trajectory $z^*_{0:T}$}
    \Notation{$\mathbb{P}(\cdot)$ denotes a probability mass function.}
    \tcc{Forward filtering}
    Set $z_0^1 = z_0$ and $w_0^1 = g(y_0 \given z_0)$ \;
    \For{$n \gets 2, \dots, N$}{
    Sample $z_0^n \sim \mu_\theta(\cdot)$ and $w_0^n = g(y_0 \given z_0^n)$ \;
    }
    \For{$t \gets 1, \dots, T$}{
        Set $z_t^1 = z_t$ and $w_t^1 = g(y_t \given z_t)$ \;
        \For{$n \gets 2, \dots, N$}{
            Sample $A_t^n$ with $\mathbb{P}(A_t^n = k) \propto w_{t-1}^k$ \quad \tcp{Multinomial resampling}
            Set $z_t^n \sim f_\theta(\cdot \given z_{t-1}^{A_t^n})$ and $w_t^n = g(y_t \given z_t^n)$\; \label{line:bootstrap}
        }
    }
    \tcc{Backward sampling}
    Sample $B_T$ with $\mathbb{P}(B_T = k) \propto w_T^k$ and set $z_T^* = z_T^{B_T}$ \;
    \For{$t \gets T-1, \dots, 0$}{
        Sample $B_t$ with $\mathbb{P}(B_t = k) \propto w_t^k \, p_{t+1}(z_{t+1}^{*} \given z_t^k)$ and set $z_t^* = z_t^{B_t}$ \; 
    }
    \KwRet{$z_{0:T}^*$} \;
\end{algorithm2e}
It is worth noting that the main motivation of \citet{naesseth2020markovian} in combining MSC and CSMC was to improve the behavior of CSMC, specifically in the context of improving on the \emph{bootstrap} proposal (appearing line 9 in Algorithm~\ref{alg:csmc}). 

Finally, albeit convergent, Algorithm~\ref{alg:csmc} is wasteful as it generates $T \times N$ particles and only keeps $T$ out of them. This prompted~\citet{cardoso2023state} to propose explicitly reusing all generated trajectories to reduce the variance of expectations estimates under the chain. However, they rely on a rejection-sampling method~\citep{Olsson2017Paris} which has since been proven to have infinite expected run time, making it a poor practical choice~\citep{dau2022backward}. In Algorithm~\ref{alg:rb_csmc}, we show how all particles generated may be reused to reduce the variance of the score estimator in Algorithm~\ref{alg:msc}. For $T=0$, it can be seen as a special case of \citet{schwedes2021rao} and can be extended to $T \geq 1$ by relying on the fact that it can be seen as an extension of ensemble MCMC methods~\citep{tjelmeland2004using, calderhead2014parallel} for Markovian systems~\citep[Section 2]{Finke2023csmc}. In fact, the algorithm of~\citet{cardoso2023state} can be understood as a noisy version of our Algorithm~\ref{alg:rb_csmc}.
\begin{algorithm2e}[t]
  \caption{Rao-Blackwellized CSMC}
  \label{alg:rb_csmc}
  \SetAlgoLined
  \LinesNumbered
  \DontPrintSemicolon
  \KwIn{Reference trajectory $z_{0:T}$}
  \KwOut{Estimate of the score $\widehat{\nabla_\theta \ell}(\theta)$}
  Run the forward filtering in \algorithmref{alg:csmc} to obtain the filtered particles, their weights, and the resampling indices. \;
  Sample $z^{n}_{0:T} \sim \mathcal{B}$ for all $n = 1, \dots, N$, where $\mathcal{B}$ is the backward sampler in \algorithmref{alg:csmc}. \;
  Compute $\widehat{\nabla_\theta \ell}(\theta) \gets \frac{1}{N} \sum_{n=1}^N \nabla_\theta \log p_\theta(z^{n}_{0:T}, y_{0:T})$ \;
  \Return{$\widehat{\nabla_\theta \ell}(\theta)$}
\end{algorithm2e}

\section{Numerical Evaluation}\label{sec:experiments}
To confirm the effectiveness of our proposed methodology, we evaluate it on a series of stochastic dynamical systems. We benchmark our approach against state-of-the-art data-driven algorithms from the reinforcement learning literature. Given the \emph{on-policy} nature of our method, we choose to compare with algorithms in the same class, namely trust-region policy optimization (TRPO) \citep{schulman2015trust} and proximal policy optimization (PPO) \citep{schulman2017proximal}, both implemented within the \emph{MushroomRL} framework \citep{mushroom2021github}. We also provide a comparison to a biased version of our method that estimates the score function via particle smoothing instead of CSMC. This corresponds to the expectation-maximization algorithm of \citet{kantas2015particle}\footnote{A public repository is available at \href{https://github.com/hanyas/psoc}{https://github.com/hanyas/psoc}}.

While our methodology shares similarities with reinforcement learning algorithms, a direct comparison is only valid within a certain scope. It is crucial to emphasize that our approach necessitates complete access to the stochastic dynamics, leading to a higher overall number of interactions with the underlying system during the sampling procedure in Algorithm~\ref{alg:rb_csmc}. Nonetheless, to ensure a fair qualitative comparison in the upcoming experiments, we equalize the number of system interactions accessible to all algorithms during the policy update stage, despite potential divergences in the sampling phase.

All dynamical systems involved in this evaluation are stochastic, under-actuated, and action-limited, making the control and stabilization tasks challenging. The action limits are satisfied by composing a neural Gaussian policy $\pi_{\theta}(u_{t} \given x_{t})$ with a normalizing flow that squashes, scales, and shifts the output of $\pi_{\theta}$ to lie within the bounded action domain:
\begin{equation}
    s_{t} \sim \mathcal{N}\bigl(\text{NN}(x_{t} ; \theta), \Sigma(\theta) \bigr), \quad  u_{t} = a \cdot \tanh{(s_{t})} + b,
\end{equation}
where $\text{NN}$ denotes a neural network parameterized by $\theta$ and $(a, b)$ are constants that vary depending on the system. This transformation leads to a non-Gaussian joint transition density $f_{\theta}(z_{t+1} \given z_{t})$ with finite support, highlighting the drawbacks of \emph{enabling approximations}~\citep{sarkka2023bayesian}.

\textbf{Pendulum:} This environment models a simple pendulum with unit mass and pole length. The state of the pendulum, described by its angular position and velocity, is two-dimensional. Meanwhile, the action, representing the torque, is one-dimensional. The control objective requires swinging the pendulum from its initial resting position $x_{0} = \{0, 0\}$ to the upright position $x_{g} = \{\pi, 0\}$ and stabilizing it around this inherently unstable equilibrium. The challenge lies in executing the swing-up within a fixed horizon while adhering to specified torque limits.

\textbf{Cart-Pole:} Here we model a cart-pole system. The state of the system is four-dimensional, comprising the cart's position and velocity, as well as the angle and angular velocity of the pole. The action space is one-dimensional, representing the force applied to the cart. The control objective involves swinging the pole up and balancing it by applying suitable forces. The starting configuration is defined as $x_{0} = \{0, 0, 0, 0\}$, where the cart is centered, and the pole is in a down-right position. The target state is $x_{g} = \{0, \pi, 0, 0\}$. The system is again force-limited, requiring a non-trivial optimal policy.

\textbf{Double Pendulum:} This environment simulates a double pendulum system characterized by two connected pendulum arms. The state is four-dimensional, encompassing the angular positions and velocities of both joints. The action space is two-dimensional, representing torques applied to each joint. The control task involves coordinating the motion of the double pendulum from an initial state $x_{0} = \{0, 0, 0, 0\}$ to a target state $x_{g} = \{\pi, 0, 0, 0\}$ and stabilizing it around this sensitive equilibrium. Achieving the swing-up within a specified horizon, while respecting torque constraints is especially challenging due to the chaotic nature of the dynamics.

The transition dynamics of all environments are defined by their respective ordinary differential equations \citep{underactuated}, discretized using an explicit Euler method with a step size of $dt = 0.05~\si{\second}$, and simulated for a horizon $T=100$. We augment the transition dynamics with diagonal Gaussian noise with a standard deviation of $\sigma = 0.01$.

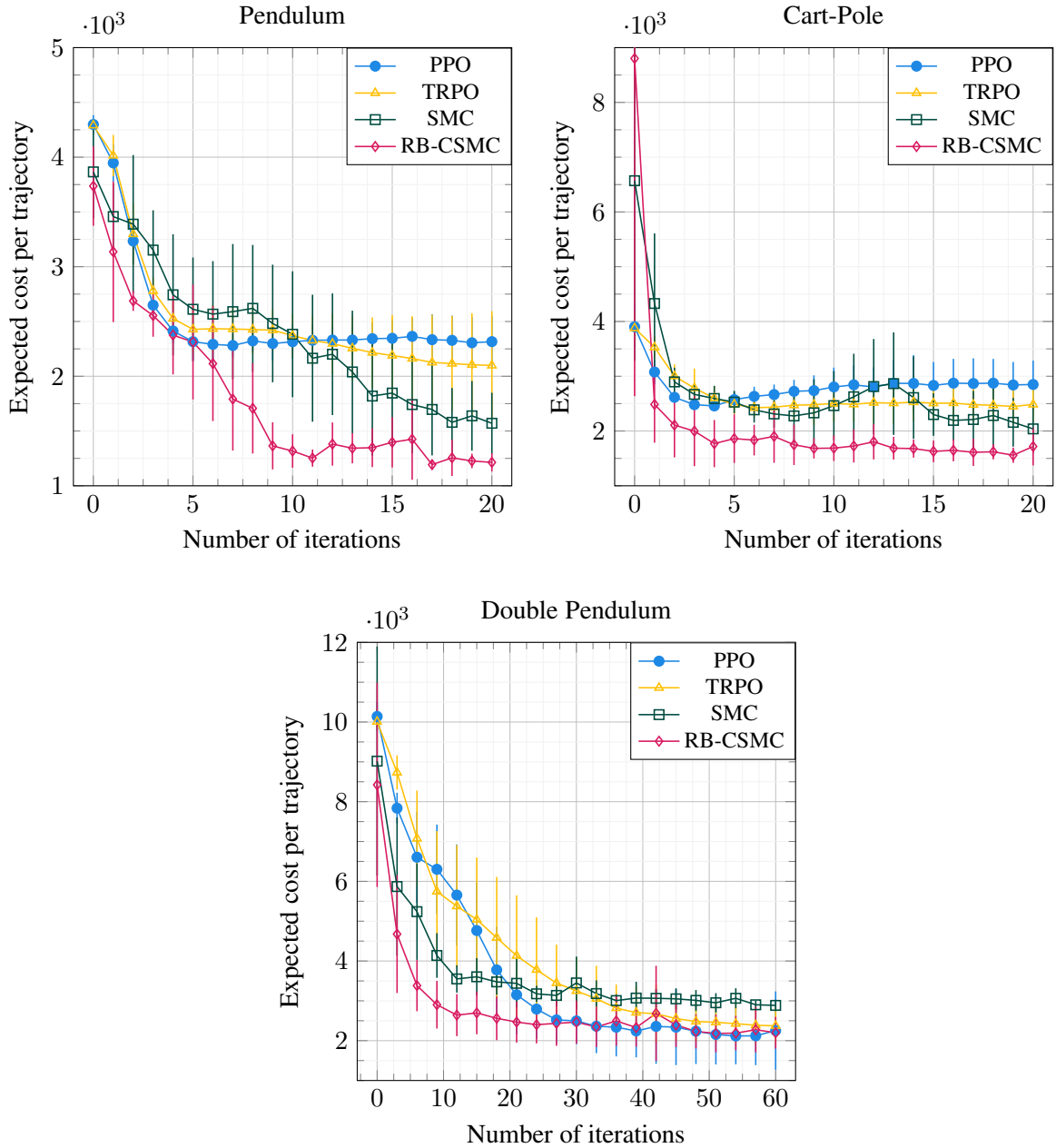
\begin{figure}[p!]
    \begin{minipage}{0.48\textwidth}
        \input{figures/pendulum.tex}
    \end{minipage}\hfill
    \begin{minipage}{0.48\textwidth}
        \input{figures/cartpole.tex}
    \end{minipage}\hfill
    \centering
    \begin{minipage}{0.48\textwidth}
        \input{figures/double_pendulum.tex}
    \end{minipage}
    \caption{Training results on the various environments. Each iteration comprises gradient updates evaluated over 15 thousand interactions with the respective environment. At every iteration, the trained policy is evaluated by averaging over 30 trajectories. We report the mean and standard deviation of the expected total cost computed over 10 random seeds. The MSC algorithm using the RB-CSMC kernel displays consistently better results than its SMC counterpart, and both perform competitively to TRPO and PPO.}
    \label{fig:algo_comparison}
\end{figure}

\figureref{fig:algo_comparison} offers a qualitative comparison between our algorithm (RB-CSMC), its particle smoothing counterpart (SMC), PPO, and TRPO on all environments. The illustrated learning curves report the mean and standard deviation of the expected total cost averaged over 10 random seeds, where each iteration accounts for 15 thousand interactions with the underlying environment. \appendixref{sec:exp_details} contains a detailed list of relevant hyperparameters.

\section{Discussion and Limitations}
We have presented a new approach to risk-sensitive stochastic optimal control within a control-as-inference framework. Our methodology relies entirely on an inference-centric paradigm and leverages state-of-the-art techniques from sequential Monte Carlo and Markov chain Monte Carlo, allowing us to tackle stochastic optimal control in a general setting without resorting to approximations and without explicit value function approximation. The reported performance is at least comparable to related approaches and seemingly validates the underlying statistical theory. 

Nonetheless, this approach is not without limitations, as it requires full access to the stochastic dynamics for sampling and log-probability evaluation, thus incurring a higher cost of interaction with the system compared to other data-driven approaches. Several open questions remain regarding a principled approach to selecting the inverse temperature parameter, addressing the infinite-horizon objective, and integrating online model inference.

\section{Contribution Statement}
The original idea is to due to HA and was further developed in conversations with SI and AC. The implementation and empirical evaluation were lead by HA with help from SI. AC advised on the underlying statistical theory and its implementation. HA, SI, and AC wrote the manuscript. SS provided regular supervision and reviewed the manuscript.

\appendix
\section{Experiment Details}\label{sec:exp_details}

\begin{table}[h]
    \centering
    \begin{tabular}{ccccc}
        \toprule
         Algorithm & Hyperparameters & Pendulum & Cart-Pole & Double Pendulum \\
         \midrule
         \multirow{5}{*}{RB-CSMC} & Tempering ($\eta$) & \num{5e-2} & \num{e-1} & \num{5e-3} \\
         & Learning rate & \num{e-2} & \num{e-3} & \num{5e-4} \\
         & Forward particles & 256 & 256 & 512 \\
         & Backward samples & 30 & 30 & 30 \\
         & Policy NN layers & 256, 256 &  256, 256 &  256, 256 \\
         \midrule
         \multirow{5}{*}{SMC} & Tempering ($\eta$) & \num{5e-2} & \num{e-1} & \num{5e-3} \\
         & Learning rate & \num{e-2} & \num{e-3} & \num{5e-4} \\
         & Forward particles & 256 & 256 & 512 \\
         & Backward samples & 30 & 30 & 30 \\
         & Policy NN layers & 256, 256 &  256, 256 &  256, 256 \\
         \bottomrule
    \end{tabular}
    \caption{Hyperparameter specifications for SMC and RB-CSMC algorithms.}
    \label{tab:csmc_parameters}
\end{table}

\vspace{-0.5cm}

\begin{table}[h]
    \centering
    \begin{tabular}{ccccc}
        \toprule
         Algorithm & Hyperparameters & Pendulum & Cart-Pole & Double Pendulum \\
         \midrule
         \multirow{3}{*}{TRPO} & KL bound & \num{e-2} & \num{5e-3} & \num{5e-3} \\
         & Critic NN layers & 256, 256 & 256, 256 & 256, 256 \\
         & Policy NN layers & 256, 256 & 256, 256 & 256, 256 \\
         & Learning rate & \num{3e-4} & \num{3e-4} & \num{3e-4} \\ \midrule
         \multirow{3}{*}{PPO} & Clipping ratio & 0.1 & 0.1 & 0.1 \\
         & Critic NN layers & 256, 256 & 256, 256 & 256, 256 \\
         & Policy NN layers & 256, 256 & 256, 256 & 256, 256 \\
         & Learning rate & \num{3e-4} & \num{3e-4} & \num{3e-4} \\
         \bottomrule
    \end{tabular}
    \caption{Hyperparameter specifications for TRPO and PPO.}
    \label{tab:ppo_parameters}
\end{table}

\newpage

\section*{Acknowledgements}
The authors express their gratitude to Joe Watson for his insightful early comments and valuable literature references. Furthermore, the authors extend their sincere appreciation to Hai-Dang Dau for his detailed comments on backward sampling techniques.

\bibliography{references.bib}

\end{document}

%% file: figures/dgm.tex
\tikzstyle{input}=[
    circle,
    very thick,
    minimum size=0.95cm,
    draw=black!80,
    fill=white!20
]

\tikzstyle{output}=[
    circle,
    very thick,
    minimum size=0.95cm,
    draw=black!80,
    fill=gray!20
]

\begin{tikzpicture}[>=latex,text height=0.8ex,text depth=0.25ex]
	\matrix[row sep=0.35cm,column sep=0.45cm] {
		& %
		\node (y_0) [output]{\large $y_0$}; &   &
		\node (y_t) [output]{\large $y_t$}; &   &
		\node (y_T) [output]{\large $y_T$}; &   &
		\\ & %
		\node (x_0) [input]{\large $x_0$}; &   &
		\node (x_t) [input]{\large $x_t$}; &   &
		\node (x_T) [input]{\large $x_T$}; &   &
		\\ & %
		\node (u_0) [input]{\large $u_0$}; &   &
		\node (u_t) [input]{\large $u_t$}; &   &
		\node (u_T) [input]{\large $u_T$}; &   &
		\\
        & \node (theta) [input]{\large $\theta$}; & \\
	};
	\path[->]
	(x_0) edge[very thick, dashed] (x_t)
	(x_t) edge[very thick, dashed] (x_T)

	(x_0) edge[very thick] (y_0)
	(x_t) edge[very thick] (y_t)
	(x_T) edge[very thick] (y_T)

	(x_0) edge[very thick] (u_0)
	(x_t) edge[very thick] (u_t)
	(x_T) edge[very thick] (u_T)

	(u_0) edge[bend right=40, very thick] (y_0)
	
	(u_t) edge[bend right=40, very thick] (y_t)
	(u_T) edge[bend right=40, very thick] (y_T)

	(u_0) edge[very thick, dashed] (x_t)
	(u_t) edge[very thick, dashed] (x_T)

	(theta) edge[very thick] (u_0)
	(theta) edge[bend right=20, very thick] (u_t)
	(theta) edge[bend right=20, very thick] (u_T)
	;
\end{tikzpicture}

%% file: figures/pendulum.tex
\begin{tikzpicture}

\definecolor{crimson2143940}{RGB}{216,27,96}
\definecolor{darkgray176}{RGB}{176,176,176}
\definecolor{darkorange25512714}{RGB}{255,193,7}
\definecolor{forestgreen4416044}{RGB}{0,77,64}
\definecolor{steelblue31119180}{RGB}{30,136,229}

\begin{axis}[
    width=8cm,
    height=8cm,
    scaled y ticks=base 10:-3,
    tick pos=both,
    grid=both,
    minor tick num=3,
    try min ticks=6,
    grid style={line width=.1pt, draw=gray!10},
    major grid style={line width=.1pt, draw=gray!50},
    xmin=-1, xmax=21,
    ymin=1000.0, ymax=5000,
    legend style={
        nodes={scale=0.85, transform shape},
        at={(1,1)},
        anchor=north east
    },
    xlabel=Number of iterations,
    ylabel=Expected cost per trajectory,
    title=Pendulum,
]
\path [draw=steelblue31119180, semithick]
(axis cs:0,4213.08404695505)
--(axis cs:0,4383.76933169296);

\path [draw=steelblue31119180, semithick]
(axis cs:1,3775.56287002511)
--(axis cs:1,4119.62801186385);

\path [draw=steelblue31119180, semithick]
(axis cs:2,3022.29327301415)
--(axis cs:2,3446.61464531055);

\path [draw=steelblue31119180, semithick]
(axis cs:3,2517.23437258852)
--(axis cs:3,2780.80427615166);

\path [draw=steelblue31119180, semithick]
(axis cs:4,2299.88077296347)
--(axis cs:4,2525.94928031385);

\path [draw=steelblue31119180, semithick]
(axis cs:5,2138.32954347793)
--(axis cs:5,2492.62625290543);

\path [draw=steelblue31119180, semithick]
(axis cs:6,2159.27956421855)
--(axis cs:6,2423.00547662704);

\path [draw=steelblue31119180, semithick]
(axis cs:7,2174.94749330853)
--(axis cs:7,2386.14704406485);

\path [draw=steelblue31119180, semithick]
(axis cs:8,2225.43660736643)
--(axis cs:8,2419.84839028317);

\path [draw=steelblue31119180, semithick]
(axis cs:9,2143.64884788487)
--(axis cs:9,2453.15997325077);

\path [draw=steelblue31119180, semithick]
(axis cs:10,2159.56346284893)
--(axis cs:10,2470.14376272983);

\path [draw=steelblue31119180, semithick]
(axis cs:11,2178.90085494895)
--(axis cs:11,2477.54482120343);

\path [draw=steelblue31119180, semithick]
(axis cs:12,2203.94642704769)
--(axis cs:12,2457.00172031731);

\path [draw=steelblue31119180, semithick]
(axis cs:13,2211.1149699711)
--(axis cs:13,2450.3239529049);

\path [draw=steelblue31119180, semithick]
(axis cs:14,2195.72536981794)
--(axis cs:14,2490.61066381945);

\path [draw=steelblue31119180, semithick]
(axis cs:15,2179.68376250537)
--(axis cs:15,2512.96021049939);

\path [draw=steelblue31119180, semithick]
(axis cs:16,2195.01658585584)
--(axis cs:16,2533.06720303417);

\path [draw=steelblue31119180, semithick]
(axis cs:17,2102.74294389463)
--(axis cs:17,2565.08083586603);

\path [draw=steelblue31119180, semithick]
(axis cs:18,2102.54280553404)
--(axis cs:18,2553.16767707172);

\path [draw=steelblue31119180, semithick]
(axis cs:19,2067.6299288086)
--(axis cs:19,2543.60248912935);

\path [draw=steelblue31119180, semithick]
(axis cs:20,2081.18600684059)
--(axis cs:20,2550.47187916008);

\path [draw=darkorange25512714, semithick]
(axis cs:0,4239.98124787855)
--(axis cs:0,4340.69884211293);

\path [draw=darkorange25512714, semithick]
(axis cs:1,3826.74864695812)
--(axis cs:1,4203.06402355079);

\path [draw=darkorange25512714, semithick]
(axis cs:2,3015.36232947868)
--(axis cs:2,3580.4074509037);

\path [draw=darkorange25512714, semithick]
(axis cs:3,2654.28831213268)
--(axis cs:3,2904.00025846676);

\path [draw=darkorange25512714, semithick]
(axis cs:4,2355.77001486504)
--(axis cs:4,2697.88825899613);

\path [draw=darkorange25512714, semithick]
(axis cs:5,2214.33603652361)
--(axis cs:5,2641.8426610932);

\path [draw=darkorange25512714, semithick]
(axis cs:6,2217.03531670093)
--(axis cs:6,2649.98569125761);

\path [draw=darkorange25512714, semithick]
(axis cs:7,2226.57151783466)
--(axis cs:7,2635.95837111316);

\path [draw=darkorange25512714, semithick]
(axis cs:8,2204.78574666052)
--(axis cs:8,2647.32722715919);

\path [draw=darkorange25512714, semithick]
(axis cs:9,2231.70351369988)
--(axis cs:9,2610.70479530196);

\path [draw=darkorange25512714, semithick]
(axis cs:10,2168.8311869467)
--(axis cs:10,2572.56343515938);

\path [draw=darkorange25512714, semithick]
(axis cs:11,2097.74123188169)
--(axis cs:11,2555.62690127686);

\path [draw=darkorange25512714, semithick]
(axis cs:12,2019.13744825273)
--(axis cs:12,2576.63077460779);

\path [draw=darkorange25512714, semithick]
(axis cs:13,1963.09300703378)
--(axis cs:13,2546.50465193744);

\path [draw=darkorange25512714, semithick]
(axis cs:14,1894.90643406195)
--(axis cs:14,2536.74853758241);

\path [draw=darkorange25512714, semithick]
(axis cs:15,1821.52539871132)
--(axis cs:15,2557.36804466003);

\path [draw=darkorange25512714, semithick]
(axis cs:16,1775.22251477487)
--(axis cs:16,2548.30977574898);

\path [draw=darkorange25512714, semithick]
(axis cs:17,1691.72585215407)
--(axis cs:17,2560.67936098241);

\path [draw=darkorange25512714, semithick]
(axis cs:18,1680.54674997795)
--(axis cs:18,2550.83208590326);

\path [draw=darkorange25512714, semithick]
(axis cs:19,1635.20072117101)
--(axis cs:19,2574.32753796441);

\path [draw=darkorange25512714, semithick]
(axis cs:20,1603.30702924296)
--(axis cs:20,2593.7196781058);

\path [draw=forestgreen4416044, semithick]
(axis cs:0,3442.01572978692)
--(axis cs:0,4289.16582501954);

\path [draw=forestgreen4416044, semithick]
(axis cs:1,3156.99987110572)
--(axis cs:1,3758.39806642996);

\path [draw=forestgreen4416044, semithick]
(axis cs:2,2757.60363474025)
--(axis cs:2,4019.43206326942);

\path [draw=forestgreen4416044, semithick]
(axis cs:3,2788.22611800158)
--(axis cs:3,3515.35309762942);

\path [draw=forestgreen4416044, semithick]
(axis cs:4,2190.67683249995)
--(axis cs:4,3296.1889484115);

\path [draw=forestgreen4416044, semithick]
(axis cs:5,2138.21432312689)
--(axis cs:5,3084.55039735666);

\path [draw=forestgreen4416044, semithick]
(axis cs:6,2083.69040654514)
--(axis cs:6,3050.48580389767);

\path [draw=forestgreen4416044, semithick]
(axis cs:7,1973.00114445724)
--(axis cs:7,3206.83644304289);

\path [draw=forestgreen4416044, semithick]
(axis cs:8,2040.92135613399)
--(axis cs:8,3198.7055170687);

\path [draw=forestgreen4416044, semithick]
(axis cs:9,1944.92215987579)
--(axis cs:9,3018.36426528633);

\path [draw=forestgreen4416044, semithick]
(axis cs:10,1809.09504295236)
--(axis cs:10,2957.80856037678);

\path [draw=forestgreen4416044, semithick]
(axis cs:11,1584.94139443545)
--(axis cs:11,2744.26007916418);

\path [draw=forestgreen4416044, semithick]
(axis cs:12,1645.37521835516)
--(axis cs:12,2756.337980332);

\path [draw=forestgreen4416044, semithick]
(axis cs:13,1480.00841059511)
--(axis cs:13,2598.6146366905);

\path [draw=forestgreen4416044, semithick]
(axis cs:14,1275.46615209839)
--(axis cs:14,2362.84073254874);

\path [draw=forestgreen4416044, semithick]
(axis cs:15,1346.35172396132)
--(axis cs:15,2346.37625327005);

\path [draw=forestgreen4416044, semithick]
(axis cs:16,1297.93533328911)
--(axis cs:16,2185.10033150878);

\path [draw=forestgreen4416044, semithick]
(axis cs:17,1277.19002973573)
--(axis cs:17,2116.07544398307);

\path [draw=forestgreen4416044, semithick]
(axis cs:18,1266.56941132683)
--(axis cs:18,1892.39760237448);

\path [draw=forestgreen4416044, semithick]
(axis cs:19,1320.60228106896)
--(axis cs:19,1956.36435440566);

\path [draw=forestgreen4416044, semithick]
(axis cs:20,1292.33586897423)
--(axis cs:20,1849.15478148262);

\path [draw=crimson2143940, semithick]
(axis cs:0,3374.04422651336)
--(axis cs:0,4098.28744832855);

\path [draw=crimson2143940, semithick]
(axis cs:1,2494.19573712226)
--(axis cs:1,3778.82167269863);

\path [draw=crimson2143940, semithick]
(axis cs:2,2596.36304924003)
--(axis cs:2,2778.94166977038);

\path [draw=crimson2143940, semithick]
(axis cs:3,2361.00115644405)
--(axis cs:3,2746.03913560285);

\path [draw=crimson2143940, semithick]
(axis cs:4,2016.70231027374)
--(axis cs:4,2734.47842958715);

\path [draw=crimson2143940, semithick]
(axis cs:5,1788.2349421928)
--(axis cs:5,2837.20603846252);

\path [draw=crimson2143940, semithick]
(axis cs:6,1590.77579784608)
--(axis cs:6,2639.04962982006);

\path [draw=crimson2143940, semithick]
(axis cs:7,1321.95010567279)
--(axis cs:7,2261.33483333028);

\path [draw=crimson2143940, semithick]
(axis cs:8,1294.70739975606)
--(axis cs:8,2118.8970732481);

\path [draw=crimson2143940, semithick]
(axis cs:9,1149.41806636271)
--(axis cs:9,1579.40247124252);

\path [draw=crimson2143940, semithick]
(axis cs:10,1164.70885067031)
--(axis cs:10,1469.57796944479);

\path [draw=crimson2143940, semithick]
(axis cs:11,1174.13071050182)
--(axis cs:11,1331.37310082915);

\path [draw=crimson2143940, semithick]
(axis cs:12,1184.17889536996)
--(axis cs:12,1577.87530063994);

\path [draw=crimson2143940, semithick]
(axis cs:13,1205.0299803352)
--(axis cs:13,1480.78935522216);

\path [draw=crimson2143940, semithick]
(axis cs:14,1170.74086009759)
--(axis cs:14,1524.53701744321);

\path [draw=crimson2143940, semithick]
(axis cs:15,1167.12486776988)
--(axis cs:15,1626.89450171732);

\path [draw=crimson2143940, semithick]
(axis cs:16,1054.57178623361)
--(axis cs:16,1796.06619572116);

\path [draw=crimson2143940, semithick]
(axis cs:17,1157.60542369134)
--(axis cs:17,1228.52407835813);

\path [draw=crimson2143940, semithick]
(axis cs:18,1089.69272446918)
--(axis cs:18,1420.72100222113);

\path [draw=crimson2143940, semithick]
(axis cs:19,1160.48002628421)
--(axis cs:19,1292.62191616333);

\path [draw=crimson2143940, semithick]
(axis cs:20,1130.14499141678)
--(axis cs:20,1297.28869719049);

\addplot [semithick, steelblue31119180, mark=*, mark size=2, mark options={solid}]
table {%
0 4298.426689324
1 3947.59544094448
2 3234.45395916235
3 2649.01932437009
4 2412.91502663866
5 2315.47789819168
6 2291.1425204228
7 2280.54726868669
8 2322.6424988248
9 2298.40441056782
10 2314.85361278938
11 2328.22283807619
12 2330.4740736825
13 2330.719461438
14 2343.16801681869
15 2346.32198650238
16 2364.041894445
17 2333.91188988033
18 2327.85524130288
19 2305.61620896898
20 2315.82894300034
};
\addplot [semithick, darkorange25512714, mark=triangle, mark size=2, mark options={solid}]
table {%
0 4290.34004499574
1 4014.90633525445
2 3297.88489019119
3 2779.14428529972
4 2526.82913693059
5 2428.08934880841
6 2433.51050397927
7 2431.26494447391
8 2426.05648690985
9 2421.20415450092
10 2370.69731105304
11 2326.68406657927
12 2297.88411143026
13 2254.79882948561
14 2215.82748582218
15 2189.44672168567
16 2161.76614526193
17 2126.20260656824
18 2115.68941794061
19 2104.76412956771
20 2098.51335367438
};
\addplot [semithick, forestgreen4416044, mark=square, mark size=2, mark options={solid}]
table {%
0 3865.59077740323
1 3457.69896876784
2 3388.51784900483
3 3151.7896078155
4 2743.43289045573
5 2611.38236024177
6 2567.0881052214
7 2589.91879375007
8 2619.81343660135
9 2481.64321258106
10 2383.45180166457
11 2164.60073679982
12 2200.85659934358
13 2039.3115236428
14 1819.15344232357
15 1846.36398861569
16 1741.51783239895
17 1696.6327368594
18 1579.48350685065
19 1638.48331773731
20 1570.74532522843
};
\addplot [semithick, crimson2143940, mark=diamond, mark size=2, mark options={solid}]
table {%
0 3736.16583742096
1 3136.50870491045
2 2687.6523595052
3 2553.52014602345
4 2375.59036993045
5 2312.72049032766
6 2114.91271383307
7 1791.64246950153
8 1706.80223650208
9 1364.41026880261
10 1317.14341005755
11 1252.75190566548
12 1381.02709800495
13 1342.90966777868
14 1347.6389387704
15 1397.0096847436
16 1425.31899097738
17 1193.06475102473
18 1255.20686334516
19 1226.55097122377
20 1213.71684430363
};
\legend{PPO, TRPO, SMC, RB-CSMC}
\end{axis}

\end{tikzpicture}

%% file: figures/cartpole.tex
\begin{tikzpicture}

\definecolor{crimson2143940}{RGB}{216,27,96}
\definecolor{darkgray176}{RGB}{176,176,176}
\definecolor{darkorange25512714}{RGB}{255,193,7}
\definecolor{forestgreen4416044}{RGB}{0,77,64}
\definecolor{steelblue31119180}{RGB}{30,136,229}

\begin{axis}[
    width=8cm,
    height=8cm,
    scaled y ticks=base 10:-3,
    tick pos=both,
    grid=both,
    minor tick num=3,
    try min ticks=6,
    grid style={line width=.1pt, draw=gray!10},
    major grid style={line width=.1pt, draw=gray!50},
    xmin=-1, xmax=21,
    ymin=1000, ymax=9000,
    legend style={
        nodes={scale=0.85, transform shape},
        at={(1,1)},
        anchor=north east
    },
    xlabel=Number of iterations,
    ylabel=Expected cost per trajectory,
    title=Cart-Pole,
]
\path [draw=steelblue31119180, semithick]
(axis cs:0,3514.59883405117)
--(axis cs:0,4292.48191913703);

\path [draw=steelblue31119180, semithick]
(axis cs:1,2856.16148572971)
--(axis cs:1,3298.18925469358);

\path [draw=steelblue31119180, semithick]
(axis cs:2,2557.9794151522)
--(axis cs:2,2672.09574269788);

\path [draw=steelblue31119180, semithick]
(axis cs:3,2415.62720876809)
--(axis cs:3,2546.65808905772);

\path [draw=steelblue31119180, semithick]
(axis cs:4,2366.65596365858)
--(axis cs:4,2548.60424634038);

\path [draw=steelblue31119180, semithick]
(axis cs:5,2427.04795264214)
--(axis cs:5,2719.24432191694);

\path [draw=steelblue31119180, semithick]
(axis cs:6,2457.01369302183)
--(axis cs:6,2807.91746193022);

\path [draw=steelblue31119180, semithick]
(axis cs:7,2482.76694051297)
--(axis cs:7,2848.10069135146);

\path [draw=steelblue31119180, semithick]
(axis cs:8,2513.37587352098)
--(axis cs:8,2935.05047909243);

\path [draw=steelblue31119180, semithick]
(axis cs:9,2458.71210860727)
--(axis cs:9,3016.66784094755);

\path [draw=steelblue31119180, semithick]
(axis cs:10,2449.85071445921)
--(axis cs:10,3157.58738699016);

\path [draw=steelblue31119180, semithick]
(axis cs:11,2486.34671274944)
--(axis cs:11,3199.98049231875);

\path [draw=steelblue31119180, semithick]
(axis cs:12,2443.91162558427)
--(axis cs:12,3162.42395653683);

\path [draw=steelblue31119180, semithick]
(axis cs:13,2441.72308714191)
--(axis cs:13,3300.72298415089);

\path [draw=steelblue31119180, semithick]
(axis cs:14,2352.70554629693)
--(axis cs:14,3385.33542101324);

\path [draw=steelblue31119180, semithick]
(axis cs:15,2404.66412054652)
--(axis cs:15,3260.18181781051);

\path [draw=steelblue31119180, semithick]
(axis cs:16,2430.88316854844)
--(axis cs:16,3317.3816868091);

\path [draw=steelblue31119180, semithick]
(axis cs:17,2407.84979804146)
--(axis cs:17,3327.45761345584);

\path [draw=steelblue31119180, semithick]
(axis cs:18,2431.84925815291)
--(axis cs:18,3316.24252705514);

\path [draw=steelblue31119180, semithick]
(axis cs:19,2421.61828318772)
--(axis cs:19,3259.94380817981);

\path [draw=steelblue31119180, semithick]
(axis cs:20,2414.01562320355)
--(axis cs:20,3285.67022857181);

\path [draw=darkorange25512714, semithick]
(axis cs:0,3595.67863644136)
--(axis cs:0,4161.08304079065);

\path [draw=darkorange25512714, semithick]
(axis cs:1,3325.65233124313)
--(axis cs:1,3733.03959686048);

\path [draw=darkorange25512714, semithick]
(axis cs:2,2787.21592614498)
--(axis cs:2,3217.057035931);

\path [draw=darkorange25512714, semithick]
(axis cs:3,2427.0623926601)
--(axis cs:3,3139.59168518089);

\path [draw=darkorange25512714, semithick]
(axis cs:4,2376.92313906172)
--(axis cs:4,2820.94729987976);

\path [draw=darkorange25512714, semithick]
(axis cs:5,2305.34504119129)
--(axis cs:5,2672.9300003075);

\path [draw=darkorange25512714, semithick]
(axis cs:6,2261.14019556055)
--(axis cs:6,2588.28420338697);

\path [draw=darkorange25512714, semithick]
(axis cs:7,2248.40612127092)
--(axis cs:7,2623.08930437373);

\path [draw=darkorange25512714, semithick]
(axis cs:8,2191.28666196223)
--(axis cs:8,2749.23314605295);

\path [draw=darkorange25512714, semithick]
(axis cs:9,2109.97967648965)
--(axis cs:9,2850.45686630929);

\path [draw=darkorange25512714, semithick]
(axis cs:10,2162.9101350743)
--(axis cs:10,2841.43213094229);

\path [draw=darkorange25512714, semithick]
(axis cs:11,2165.89935412759)
--(axis cs:11,2809.28287617367);

\path [draw=darkorange25512714, semithick]
(axis cs:12,2138.81605009428)
--(axis cs:12,2895.34106421726);

\path [draw=darkorange25512714, semithick]
(axis cs:13,2134.71651274591)
--(axis cs:13,2882.7258019441);

\path [draw=darkorange25512714, semithick]
(axis cs:14,2184.70860586695)
--(axis cs:14,2871.49304960635);

\path [draw=darkorange25512714, semithick]
(axis cs:15,2175.91911700133)
--(axis cs:15,2837.98662439978);

\path [draw=darkorange25512714, semithick]
(axis cs:16,2140.01961229716)
--(axis cs:16,2880.53007456611);

\path [draw=darkorange25512714, semithick]
(axis cs:17,2167.11551615577)
--(axis cs:17,2793.64615801525);

\path [draw=darkorange25512714, semithick]
(axis cs:18,2168.78338415491)
--(axis cs:18,2772.20997223446);

\path [draw=darkorange25512714, semithick]
(axis cs:19,2120.19688096146)
--(axis cs:19,2777.00589541093);

\path [draw=darkorange25512714, semithick]
(axis cs:20,2164.72757891378)
--(axis cs:20,2804.04238342474);

\path [draw=forestgreen4416044, semithick]
(axis cs:0,3286.44725207109)
--(axis cs:0,9854.01256483686);

\path [draw=forestgreen4416044, semithick]
(axis cs:1,3048.82372814697)
--(axis cs:1,5607.77322351749);

\path [draw=forestgreen4416044, semithick]
(axis cs:2,2637.64485233356)
--(axis cs:2,3155.66094493643);

\path [draw=forestgreen4416044, semithick]
(axis cs:3,2467.85475236433)
--(axis cs:3,2869.42322538535);

\path [draw=forestgreen4416044, semithick]
(axis cs:4,2361.84949623607)
--(axis cs:4,2825.42065168267);

\path [draw=forestgreen4416044, semithick]
(axis cs:5,2322.57882569076)
--(axis cs:5,2733.26657578471);

\path [draw=forestgreen4416044, semithick]
(axis cs:6,2154.69355889984)
--(axis cs:6,2619.18388610438);

\path [draw=forestgreen4416044, semithick]
(axis cs:7,2061.92009127385)
--(axis cs:7,2561.44874072853);

\path [draw=forestgreen4416044, semithick]
(axis cs:8,1782.7945459782)
--(axis cs:8,2767.4801816319);

\path [draw=forestgreen4416044, semithick]
(axis cs:9,1716.96710114668)
--(axis cs:9,2946.22642168012);

\path [draw=forestgreen4416044, semithick]
(axis cs:10,1831.24788914492)
--(axis cs:10,3095.59423766039);

\path [draw=forestgreen4416044, semithick]
(axis cs:11,1839.09081366049)
--(axis cs:11,3410.67963749115);

\path [draw=forestgreen4416044, semithick]
(axis cs:12,1943.81268087774)
--(axis cs:12,3679.51850243148);

\path [draw=forestgreen4416044, semithick]
(axis cs:13,1928.95492680777)
--(axis cs:13,3801.24701396931);

\path [draw=forestgreen4416044, semithick]
(axis cs:14,1857.49481909867)
--(axis cs:14,3362.40351006053);

\path [draw=forestgreen4416044, semithick]
(axis cs:15,1912.3039750197)
--(axis cs:15,2688.38372938353);

\path [draw=forestgreen4416044, semithick]
(axis cs:16,1827.05797758798)
--(axis cs:16,2560.77923486197);

\path [draw=forestgreen4416044, semithick]
(axis cs:17,1687.90461649693)
--(axis cs:17,2732.48758453913);

\path [draw=forestgreen4416044, semithick]
(axis cs:18,1695.81900008189)
--(axis cs:18,2870.13446978204);

\path [draw=forestgreen4416044, semithick]
(axis cs:19,1714.34292450914)
--(axis cs:19,2602.83840741546);

\path [draw=forestgreen4416044, semithick]
(axis cs:20,1601.17907284307)
--(axis cs:20,2476.20979477816);

\path [draw=crimson2143940, semithick]
(axis cs:0,2636.16752100694)
--(axis cs:0,14968.5458487787);

\path [draw=crimson2143940, semithick]
(axis cs:1,1787.0565909443)
--(axis cs:1,3184.75324542095);

\path [draw=crimson2143940, semithick]
(axis cs:2,1518.81098052665)
--(axis cs:2,2696.81540980844);

\path [draw=crimson2143940, semithick]
(axis cs:3,1358.16785344453)
--(axis cs:3,2637.29468922661);

\path [draw=crimson2143940, semithick]
(axis cs:4,1339.55021279615)
--(axis cs:4,2201.31896574572);

\path [draw=crimson2143940, semithick]
(axis cs:5,1417.0502965748)
--(axis cs:5,2304.88778839532);

\path [draw=crimson2143940, semithick]
(axis cs:6,1554.41752123425)
--(axis cs:6,2113.66560075351);

\path [draw=crimson2143940, semithick]
(axis cs:7,1420.16771625023)
--(axis cs:7,2383.63203659687);

\path [draw=crimson2143940, semithick]
(axis cs:8,1379.5683562811)
--(axis cs:8,2117.52128806283);

\path [draw=crimson2143940, semithick]
(axis cs:9,1497.9234482507)
--(axis cs:9,1869.10834360893);

\path [draw=crimson2143940, semithick]
(axis cs:10,1453.01365634792)
--(axis cs:10,1921.44482964988);

\path [draw=crimson2143940, semithick]
(axis cs:11,1424.88457860972)
--(axis cs:11,2027.43241560029);

\path [draw=crimson2143940, semithick]
(axis cs:12,1482.18689238059)
--(axis cs:12,2125.17289833442);

\path [draw=crimson2143940, semithick]
(axis cs:13,1479.33915140981)
--(axis cs:13,1894.48466841584);

\path [draw=crimson2143940, semithick]
(axis cs:14,1513.96679923407)
--(axis cs:14,1844.233416284);

\path [draw=crimson2143940, semithick]
(axis cs:15,1431.3873746189)
--(axis cs:15,1825.74186921492);

\path [draw=crimson2143940, semithick]
(axis cs:16,1446.24558571531)
--(axis cs:16,1848.36774342474);

\path [draw=crimson2143940, semithick]
(axis cs:17,1363.06193250651)
--(axis cs:17,1859.68161200067);

\path [draw=crimson2143940, semithick]
(axis cs:18,1482.09925442005)
--(axis cs:18,1759.3095356151);

\path [draw=crimson2143940, semithick]
(axis cs:19,1419.64919707007)
--(axis cs:19,1696.1577228295);

\path [draw=crimson2143940, semithick]
(axis cs:20,1372.05032353848)
--(axis cs:20,2063.50788850908);

\addplot [semithick, steelblue31119180, mark=*, mark size=2, mark options={solid}]
table {%
0 3903.5403765941
1 3077.17537021165
2 2615.03757892504
3 2481.14264891291
4 2457.63010499948
5 2573.14613727954
6 2632.46557747603
7 2665.43381593222
8 2724.21317630671
9 2737.68997477741
10 2803.71905072468
11 2843.1636025341
12 2803.16779106055
13 2871.2230356464
14 2869.02048365509
15 2832.42296917852
16 2874.13242767877
17 2867.65370574865
18 2874.04589260403
19 2840.78104568376
20 2849.84292588768
};
\addplot [semithick, darkorange25512714, mark=triangle, mark size=2, mark options={solid}]
table {%
0 3878.380838616
1 3529.3459640518
2 3002.13648103799
3 2783.3270389205
4 2598.93521947074
5 2489.1375207494
6 2424.71219947376
7 2435.74771282233
8 2470.25990400759
9 2480.21827139947
10 2502.1711330083
11 2487.59111515063
12 2517.07855715577
13 2508.72115734501
14 2528.10082773665
15 2506.95287070056
16 2510.27484343164
17 2480.38083708551
18 2470.49667819469
19 2448.60138818619
20 2484.38498116926
};
\addplot [semithick, forestgreen4416044, mark=square, mark size=2, mark options={solid}]
table {%
0 6570.22990845398
1 4328.29847583223
2 2896.65289863499
3 2668.63898887484
4 2593.63507395937
5 2527.92270073774
6 2386.93872250211
7 2311.68441600119
8 2275.13736380505
9 2331.5967614134
10 2463.42106340266
11 2624.88522557582
12 2811.66559165461
13 2865.10097038854
14 2609.9491645796
15 2300.34385220162
16 2193.91860622498
17 2210.19610051803
18 2282.97673493196
19 2158.5906659623
20 2038.69443381062
};
\addplot [semithick, crimson2143940, mark=diamond, mark size=2, mark options={solid}]
table {%
0 8802.3566848928
1 2485.90491818263
2 2107.81319516754
3 1997.73127133557
4 1770.43458927094
5 1860.96904248506
6 1834.04156099388
7 1901.89987642355
8 1748.54482217196
9 1683.51589592981
10 1687.2292429989
11 1726.158497105
12 1803.67989535751
13 1686.91190991282
14 1679.10010775903
15 1628.56462191691
16 1647.30666457002
17 1611.37177225359
18 1620.70439501757
19 1557.90345994979
20 1717.77910602378
};
\legend{PPO, TRPO, SMC, RB-CSMC}
\end{axis}

\end{tikzpicture}

%% file: figures/double_pendulum.tex
\begin{tikzpicture}

\definecolor{crimson2143940}{RGB}{216,27,96}
\definecolor{darkgray176}{RGB}{176,176,176}
\definecolor{darkorange25512714}{RGB}{255,193,7}
\definecolor{forestgreen4416044}{RGB}{0,77,64}
\definecolor{steelblue31119180}{RGB}{30,136,229}

\begin{axis}[
    width=8cm,
    height=8cm,
    scaled y ticks=base 10:-3,
    tick pos=both,
    grid=both,
    minor tick num=3,
    try min ticks=6,
    grid style={line width=.1pt, draw=gray!10},
    major grid style={line width=.1pt, draw=gray!50},
    xmin=-3, xmax=63,
    ymin=1000, ymax=12000,
    legend style={
        nodes={scale=0.85, transform shape},
        at={(1,1)},
        anchor=north east
    },
    xlabel=Number of iterations,
    ylabel=Expected cost per trajectory,
    title=Double Pendulum,
]
\path [draw=steelblue31119180, semithick]
(axis cs:0,9781.01825640702)
--(axis cs:0,10497.8200322552);

\path [draw=steelblue31119180, semithick]
(axis cs:3,7451.26845255326)
--(axis cs:3,8223.86715581045);

\path [draw=steelblue31119180, semithick]
(axis cs:6,5458.43297695593)
--(axis cs:6,7750.87776143246);

\path [draw=steelblue31119180, semithick]
(axis cs:9,5179.34861082324)
--(axis cs:9,7425.58533132673);

\path [draw=steelblue31119180, semithick]
(axis cs:12,4382.88877574496)
--(axis cs:12,6926.52338527657);

\path [draw=steelblue31119180, semithick]
(axis cs:15,3558.47506390363)
--(axis cs:15,5970.65571946054);

\path [draw=steelblue31119180, semithick]
(axis cs:18,2688.72096864883)
--(axis cs:18,4864.92274487825);

\path [draw=steelblue31119180, semithick]
(axis cs:21,2158.62140298183)
--(axis cs:21,4145.4602272953);

\path [draw=steelblue31119180, semithick]
(axis cs:24,2051.60194378141)
--(axis cs:24,3535.31454516272);

\path [draw=steelblue31119180, semithick]
(axis cs:27,1930.4861896838)
--(axis cs:27,3114.35380555735);

\path [draw=steelblue31119180, semithick]
(axis cs:30,1912.32675974637)
--(axis cs:30,3080.0375457669);

\path [draw=steelblue31119180, semithick]
(axis cs:33,1686.24222118566)
--(axis cs:33,3039.73405202958);

\path [draw=steelblue31119180, semithick]
(axis cs:36,1604.99233887748)
--(axis cs:36,3070.26109597667);

\path [draw=steelblue31119180, semithick]
(axis cs:39,1580.49303181514)
--(axis cs:39,2907.56131651246);

\path [draw=steelblue31119180, semithick]
(axis cs:42,1419.9228223128)
--(axis cs:42,3295.21846932361);

\path [draw=steelblue31119180, semithick]
(axis cs:45,1387.71298513582)
--(axis cs:45,3289.43608491951);

\path [draw=steelblue31119180, semithick]
(axis cs:48,1413.6514188377)
--(axis cs:48,3060.95647825368);

\path [draw=steelblue31119180, semithick]
(axis cs:51,1403.40053953831)
--(axis cs:51,2905.67632442785);

\path [draw=steelblue31119180, semithick]
(axis cs:54,1407.87147796582)
--(axis cs:54,2828.13779338046);

\path [draw=steelblue31119180, semithick]
(axis cs:57,1384.79171209435)
--(axis cs:57,2858.86837023816);

\path [draw=steelblue31119180, semithick]
(axis cs:60,1268.4896321138)
--(axis cs:60,3238.06914658854);

\path [draw=darkorange25512714, semithick]
(axis cs:0,9839.80314800701)
--(axis cs:0,10178.2139816316);

\path [draw=darkorange25512714, semithick]
(axis cs:3,8305.29703559156)
--(axis cs:3,9161.7989170679);

\path [draw=darkorange25512714, semithick]
(axis cs:6,5869.70515768519)
--(axis cs:6,8278.59003358874);

\path [draw=darkorange25512714, semithick]
(axis cs:9,4230.69773073299)
--(axis cs:9,7259.48733525169);

\path [draw=darkorange25512714, semithick]
(axis cs:12,3838.74087468201)
--(axis cs:12,6912.13012008653);

\path [draw=darkorange25512714, semithick]
(axis cs:15,3481.05880791934)
--(axis cs:15,6598.82948827936);

\path [draw=darkorange25512714, semithick]
(axis cs:18,3060.86469183338)
--(axis cs:18,6109.30339765108);

\path [draw=darkorange25512714, semithick]
(axis cs:21,2617.26285893903)
--(axis cs:21,5647.73959612429);

\path [draw=darkorange25512714, semithick]
(axis cs:24,2462.03680326748)
--(axis cs:24,5097.47525535224);

\path [draw=darkorange25512714, semithick]
(axis cs:27,2479.11990587921)
--(axis cs:27,4412.71736780727);

\path [draw=darkorange25512714, semithick]
(axis cs:30,2415.09110025343)
--(axis cs:30,4083.5218485265);

\path [draw=darkorange25512714, semithick]
(axis cs:33,2228.1612819246)
--(axis cs:33,3879.64661252475);

\path [draw=darkorange25512714, semithick]
(axis cs:36,2227.21837504513)
--(axis cs:36,3409.89605806769);

\path [draw=darkorange25512714, semithick]
(axis cs:39,2039.90199024983)
--(axis cs:39,3371.23155846448);

\path [draw=darkorange25512714, semithick]
(axis cs:42,1945.90122774608)
--(axis cs:42,3408.62674803467);

\path [draw=darkorange25512714, semithick]
(axis cs:45,2000.76006181235)
--(axis cs:45,3108.14223029226);

\path [draw=darkorange25512714, semithick]
(axis cs:48,1967.68491920454)
--(axis cs:48,2997.56542518049);

\path [draw=darkorange25512714, semithick]
(axis cs:51,1968.54770629822)
--(axis cs:51,2952.31264974986);

\path [draw=darkorange25512714, semithick]
(axis cs:54,1927.13576046316)
--(axis cs:54,2926.61007495138);

\path [draw=darkorange25512714, semithick]
(axis cs:57,1910.50324523432)
--(axis cs:57,2862.95880668025);

\path [draw=darkorange25512714, semithick]
(axis cs:60,1906.59898609545)
--(axis cs:60,2838.40992597584);

\path [draw=forestgreen4416044, semithick]
(axis cs:0,6142.52579850615)
--(axis cs:0,11893.4519281822);

\path [draw=forestgreen4416044, semithick]
(axis cs:3,4134.79499521663)
--(axis cs:3,7604.26134392331);

\path [draw=forestgreen4416044, semithick]
(axis cs:6,4027.30291885227)
--(axis cs:6,6450.76735457472);

\path [draw=forestgreen4416044, semithick]
(axis cs:9,3578.5300892024)
--(axis cs:9,4699.44862963747);

\path [draw=forestgreen4416044, semithick]
(axis cs:12,3206.81884789466)
--(axis cs:12,3894.53640766546);

\path [draw=forestgreen4416044, semithick]
(axis cs:15,3136.29471077372)
--(axis cs:15,4071.57722712485);

\path [draw=forestgreen4416044, semithick]
(axis cs:18,3153.48608037754)
--(axis cs:18,3797.89790841298);

\path [draw=forestgreen4416044, semithick]
(axis cs:21,2829.9703055808)
--(axis cs:21,4048.58921326533);

\path [draw=forestgreen4416044, semithick]
(axis cs:24,2974.54422262129)
--(axis cs:24,3377.47892701299);

\path [draw=forestgreen4416044, semithick]
(axis cs:27,2861.81129707118)
--(axis cs:27,3410.59729988973);

\path [draw=forestgreen4416044, semithick]
(axis cs:30,2786.05222318068)
--(axis cs:30,4112.58039610398);

\path [draw=forestgreen4416044, semithick]
(axis cs:33,2850.28545566731)
--(axis cs:33,3514.47556801468);

\path [draw=forestgreen4416044, semithick]
(axis cs:36,2848.07868710722)
--(axis cs:36,3165.34638798955);

\path [draw=forestgreen4416044, semithick]
(axis cs:39,2660.10558749268)
--(axis cs:39,3478.97251772312);

\path [draw=forestgreen4416044, semithick]
(axis cs:42,2831.41337241338)
--(axis cs:42,3295.6900410662);

\path [draw=forestgreen4416044, semithick]
(axis cs:45,2786.18678335787)
--(axis cs:45,3314.45155281853);

\path [draw=forestgreen4416044, semithick]
(axis cs:48,2751.93663916066)
--(axis cs:48,3271.36280197407);

\path [draw=forestgreen4416044, semithick]
(axis cs:51,2711.08056810518)
--(axis cs:51,3193.20280733772);

\path [draw=forestgreen4416044, semithick]
(axis cs:54,2813.16577436311)
--(axis cs:54,3315.95022700947);

\path [draw=forestgreen4416044, semithick]
(axis cs:57,2797.88144711063)
--(axis cs:57,3004.61763784779);

\path [draw=forestgreen4416044, semithick]
(axis cs:60,2744.15604371294)
--(axis cs:60,3027.32380961875);

\path [draw=crimson2143940, semithick]
(axis cs:0,5859.57295771501)
--(axis cs:0,10985.0366435264);

\path [draw=crimson2143940, semithick]
(axis cs:3,3196.04456764029)
--(axis cs:3,6153.79784612544);

\path [draw=crimson2143940, semithick]
(axis cs:6,2740.11787551273)
--(axis cs:6,4023.34411878597);

\path [draw=crimson2143940, semithick]
(axis cs:9,2304.59889349306)
--(axis cs:9,3500.65423769762);

\path [draw=crimson2143940, semithick]
(axis cs:12,2113.72451473716)
--(axis cs:12,3170.3507610134);

\path [draw=crimson2143940, semithick]
(axis cs:15,2161.47444907919)
--(axis cs:15,3232.92911924918);

\path [draw=crimson2143940, semithick]
(axis cs:18,2013.15035849182)
--(axis cs:18,3110.04125059869);

\path [draw=crimson2143940, semithick]
(axis cs:21,1952.14715062923)
--(axis cs:21,2982.86731591744);

\path [draw=crimson2143940, semithick]
(axis cs:24,1931.3052961825)
--(axis cs:24,2872.6730001635);

\path [draw=crimson2143940, semithick]
(axis cs:27,1870.63936145294)
--(axis cs:27,2997.9085167408);

\path [draw=crimson2143940, semithick]
(axis cs:30,1923.9421503048)
--(axis cs:30,3005.82294804365);

\path [draw=crimson2143940, semithick]
(axis cs:33,1838.95775756318)
--(axis cs:33,2849.65055813548);

\path [draw=crimson2143940, semithick]
(axis cs:36,1870.6301455889)
--(axis cs:36,3120.97833319712);

\path [draw=crimson2143940, semithick]
(axis cs:39,1850.07363534892)
--(axis cs:39,2808.43573727493);

\path [draw=crimson2143940, semithick]
(axis cs:42,1491.02046305185)
--(axis cs:42,3879.8960688791);

\path [draw=crimson2143940, semithick]
(axis cs:45,1838.67762800455)
--(axis cs:45,2949.20604944276);

\path [draw=crimson2143940, semithick]
(axis cs:48,1806.35950910498)
--(axis cs:48,2657.83037223755);

\path [draw=crimson2143940, semithick]
(axis cs:51,1709.35015053629)
--(axis cs:51,2654.56413057926);

\path [draw=crimson2143940, semithick]
(axis cs:54,1759.74874764119)
--(axis cs:54,2618.81135175111);

\path [draw=crimson2143940, semithick]
(axis cs:57,1704.59258340788)
--(axis cs:57,2845.47382689874);

\path [draw=crimson2143940, semithick]
(axis cs:60,1800.35277361407)
--(axis cs:60,2597.91120921587);

\addplot [semithick, steelblue31119180, mark=*, mark size=2, mark options={solid}]
table {%
0 10139.4191443311
3 7837.56780418185
6 6604.6553691942
9 6302.46697107499
12 5654.70608051077
15 4764.56539168208
18 3776.82185676354
21 3152.04081513857
24 2793.45824447206
27 2522.41999762058
30 2496.18215275663
33 2362.98813660762
36 2337.62671742708
39 2244.0271741638
42 2357.57064581821
45 2338.57453502767
48 2237.30394854569
51 2154.53843198308
54 2118.00463567314
57 2121.83004116625
60 2253.27938935117
};
\addplot [semithick, darkorange25512714, mark=triangle, mark size=2, mark options={solid}]
table {%
0 10009.0085648193
3 8733.54797632973
6 7074.14759563696
9 5745.09253299234
12 5375.43549738427
15 5039.94414809935
18 4585.08404474223
21 4132.50122753166
24 3779.75602930986
27 3445.91863684324
30 3249.30647438997
33 3053.90394722467
36 2818.55721655641
39 2705.56677435716
42 2677.26398789037
45 2554.45114605231
48 2482.62517219251
51 2460.43017802404
54 2426.87291770727
57 2386.73102595728
60 2372.50445603565
};
\addplot [semithick, forestgreen4416044, mark=square, mark size=2, mark options={solid}]
table {%
0 9017.98886334417
3 5869.52816956997
6 5239.0351367135
9 4138.98935941994
12 3550.67762778006
15 3603.93596894929
18 3475.69199439526
21 3439.27975942306
24 3176.01157481714
27 3136.20429848045
30 3449.31630964233
33 3182.380511841
36 3006.71253754838
39 3069.5390526079
42 3063.55170673979
45 3050.3191680882
48 3011.64972056737
51 2952.14168772145
54 3064.55800068629
57 2901.24954247921
60 2885.73992666584
};
\addplot [semithick, crimson2143940, mark=diamond, mark size=2, mark options={solid}]
table {%
0 8422.30480062068
3 4674.92120688287
6 3381.73099714935
9 2902.62656559534
12 2642.03763787528
15 2697.20178416418
18 2561.59580454526
21 2467.50723327333
24 2401.989148173
27 2434.27393909687
30 2464.88254917422
33 2344.30415784933
36 2495.80423939301
39 2329.25468631192
42 2685.45826596548
45 2393.94183872366
48 2232.09494067126
51 2181.95714055777
54 2189.28004969615
57 2275.03320515331
60 2199.13199141497
};
\legend{PPO, TRPO, SMC, RB-CSMC}
\end{axis}

\end{tikzpicture}